\documentclass{article}
\usepackage{ijcai20}

\usepackage{times}
\usepackage{soul}
\usepackage{url}
\usepackage[hidelinks]{hyperref}
\usepackage[utf8]{inputenc}
\usepackage[small]{caption}
\usepackage{graphicx}
\usepackage{amsmath}
\usepackage{amsthm}
\usepackage{booktabs}
\usepackage{algorithm}
\usepackage{algorithmic}
\usepackage{fixltx2e}
\title{Multi-task Learning with Multi-head Attention for Multi-choice Reading Comprehension}

\author{
	Hui Wan
	\affiliations
	IBM Research AI\\
	\emails
	hwan@us.ibm.com
}

\begin{document}

\maketitle

\begin{abstract}
   Multiple-choice Machine Reading Comprehension (MRC) is an important and challenging Natural Language Understanding (NLU) task, in which a machine must choose the answer to a question from a set of choices, with the question placed in context of text passages or dialog. In the last a couple of years the NLU field has been revolutionized with the advent of models based on the Transformer architecture, which are pretrained on massive amounts of unsupervised data and then fine-tuned for various supervised learning NLU tasks. Transformer models have come to dominate a wide variety of leader-boards in the NLU field; in the area of MRC, the current state-of-the-art model on the DREAM dataset (see  \cite{sundream2018}) fine tunes Albert, a large pretrained Transformer-based model, and additionally combines it with an extra layer of multi-head attention between context and question-answer~\cite{zhu2020dual}. The purpose of this note is to document a new state-of-the-art result in the DREAM task, which is accomplished by, additionally, performing multi-task learning on two MRC multi-choice reading comprehension tasks (RACE and DREAM).
  
\end{abstract}

\section{Introduction}

Making a computer system understand text within context and answer questions is challenging but has attracted a lot of interest of the Artificial Intelligence community and general audience for a long time. In the recent years, many Machine Reading Comprehension (MRC) datasets have been published, with different genres and formats of the context and questions. The context could be in the form of text passages, or in the form of dialogues. The questions could be open-formed (e.g. HotPotQA~\cite{yang2018hotpotqa}), asking the system to either extract the answers as spans from the context or external knowledge, or abstract and summarize the answers; the questions could also be in the form of asking the system to choose the best answer from multiple choices. In this note we will focus on the multi-choice MRC tasks, more specifically, the DREAM task~\cite{sundream2018}.

\subsection{Datasets}

 RACE \cite{lai-etal-2017-race} is a large-scale reading comprehension dataset with more than 28,000 passages and nearly 100,000 questions. The average passage length is 322 words. Each question provides 4 answer options to choose from. The human ceiling performance is 94.5.
 
 DREAM \cite{sundream2018} is a much smaller reading comprehension dataset with more than 6,000 dialogues and over 10,000 questions. The average dialogue length is 86 words. Each question provides 3 answer options to choose from. The human ceiling performance is 98.6.

\subsection{Related Work}

Early works on the DREAM task include feature-based GBDT~\cite{sundream2018}, and FTLM~\cite{radford2018improving} which is based on the Transformer~\cite{vaswani2017attention} architecture. The top system accuracy on the DREAM leaderboard has been advanced gradually to above 90 percent, since the break-through of the text encoder in the form of large pretrained Transformer-based models (BERT~\cite{devlin-etal-2019-bert}, XLNet~\cite{conf/nips/YangDYCSL19}, RoBERTa~\cite{DBLP:journals/corr/abs-1907-11692}, Albert~\cite{Lan2020ALBERT:}).

Transfer learning is a widely used practice in machine learning (ML) that focuses on utilizing knowledge gained while solving one problem and applying it to a different but related problem. Using pretrained language models (LMs) such as ELMO~\cite{peters-etal-2018-deep} and BERT~\cite{devlin-etal-2019-bert} in down-stream tasks is an example of sequential transfer learning. On the other hand, multi-task learning, which involves learning several similar tasks simultaneously, is able to share the knowledge learned among the tasks.

On the DREAM leaderboard\footnote{https://dataset.org/dream/}, the recent top systems include RoBERTA\textsubscript{large}+MMM in~\cite{jin2019mmm} and  Albert\textsubscript{xxlarge}+DUMA in~\cite{zhu2020dual}. Both systems employ model architectures composed of a Transformer-based encoder and some matching/attention mechanism between the context and the question-answer pair. RoBERTA\textsubscript{large}+MMM in~\cite{jin2019mmm} additionally employed two stages of transfer learning: coarse-tuning with natural language inference (NLI) tasks and multi-task learning with multi-choice reading comprehension tasks.

We used Albert\textsubscript{xxlarge}+DUMA as our model architecture and did multi-task learning on top of that, as through experiments we felt it efficiently boosts the performance on top of the powerful Albert\textsubscript{xxlarge} model.

%\subsection{Transformer Models}
%Attention is all you need, Transformer, BERT, XLNet. RoBERTa, Albert
%
%
%\subsection{Transfer Learning}
%Multi-stage -- pretraining and fine tuning.
%
%Multi-task learning is another way of transferring and sharing knowledge between tasks. 
%
%
%\subsection{Machine Reading Comprehension}
%
%Briefly some prior work

\section{Methods}

The model architecture is composed of a Transformer-based encoder, a linear layer classifier, and an extra attention layer(s) in between to model reasoning between the context and the question-answer, as in \cite{zhu2020dual}. We use the pretrained Albert\textsubscript{xxlarge} as the encoder, and fine tune it during the training process. Since the DREAM dataset is small, joint training on both the DREAM task and the RACE task helps get a good boost on the DREAM task.

\subsection{Model Architecture}

% Describe overall idea of each Q and option ***

\begin{description}
	\item[Encoder]
When encoding an answer option for a question, we concatenate it with not only the question but also the context (passage for the RACE task, dialogue for the DREAM task) to form one single sequence, the parts separated by the \verb|<sep>| token. The sequence is fed through a Transformer-based encoder (Albert in our case). The sequence output of the encoder is then sent to the next part of the model.

\item[Extra Attention layer]
In the next part of the model, we use Dual Multi-head Co-Attention (DUMA) module as described in Section 4.2 of \cite{zhu2020dual}. Basically, it involves 1) splitting the output sequence from the encoder into two parts, one for the context and one for the question-answer; 2) from the two sequences, computing two attention representations, one from the context attending the question-answer, the other vice versa; 3) the two attention representations are individually mean-pooled and then concatenated together and sent to the next part of the model: the classifier.

\item[Classifier]
For each question, over all the answer options, the classifier part takes the outputs from the Extra Attention layer, and feeds them through a linear layer. The answer option with the highest probability is picked as the predicted answer. The Cross Entropy function between the ground truth and the predicted probabilities is applied to compute the loss.
\end{description}

\subsection{Multi-task Learning}
The question/answer pairs in the DREAM task have syntactic and semantic characteristics that are generally different from the text sequences that are used to pre-train Albert. Because the DREAM dataset is relatively small, it is reasonable to hypothesize that adding a larger amount of similar multi-choice MRC data in the training will be beneficial for the DREAM task.

Inspired by \cite{jin2019mmm}, we did multi-task learning on the DREAM task and the RACE task. Although the number of choices are different in both tasks, we are still able to share all the parts of the model between them. We sampled mini-batches from the DREAM and RACE datasets in proportion of the relative sizes of the two datasets.

\section{Experiment Settings}

 We used the pretrained Albert-xxlarge-v2 model as our encoder, and one layer of DUMA as in \cite{zhu2020dual}. Since we do not have the implementation details on the number of DUMA attention heads and the head size in \cite{zhu2020dual}, for re-implementation we used the setting as in Alberta-xxlarge self attention layers: 64 attention heads and each head has a dimension of 64. We used the same setting in our multi-task learning. Our codes are written based on Transformers~\footnote{https://github.com/huggingface/transformers}.

The maximum sequence length was set to 512. We used a mini-batch size of 24, and a learning rate of 1e-05. The gradient norm was clipped to 1. We adopted a linear scheduler with a weight decay of 0.01, trained the model for 5 epochs. For the multi-task learning, we used $10\%$ of the total steps as warming up, evaluated on the dev set at every 1000 steps and saved the best model on the dev set. For the single-task training on the DREAM dataset (the second last line in Table \ref{table_all_result}), we evaluated on the dev set at every 100 steps and saved the best model on the dev set.

We did not do hyper-parameter searching and only had one run of multi-task learning at the moment. For the single task training we had three runs and picked the model with the best accuracy on the dev set. All the experiments was run on four v100 GPUs in a single machine.

\section{Results}

Table~\ref{table_all_result} summarizes the experiment results. The numbers marked with $\star$ are from ~\cite{jin2019mmm}, the numbers marked with $\dagger$ are from \cite{zhu2020dual}. Note that the second last line is our implementation of Albert\textsubscript{xxlarge}+DUMA, with 64 DUMA heads of dimension 64. The multi-task learning in the last line was run with similar settings and parameters.

Compared to \cite{zhu2020dual}, our implementation of Albert\textsubscript{xxlarge}+DUMA obtained a higher accuracy on the dev set but a lower accuracy on the test set. The possible reasons could be that we did not have the exact setting such as attention head number and size, and/or randomness. Nonetheless, the model from multi-task learning had a good boost over both the numbers from \cite{zhu2020dual} and the numbers from our re-implementation of the single-task learning. This shows that although the context part in the DREAM task is in dialogue style instead of passage style in the RACE task, the DREAM task could still benefit a lot from learning together with the RACE task, because of similar domain and similar question-answer style.

% ***When have time, add another line of result for multi-task learning with different proportion of data?***

\begin{table}
	\centering
	\begin{tabular}{lll}
		\hline
		model  & dev & test \\
		\hline
		FTLM++\cite{sundream2018}   & 58.1\textsuperscript{*}  &   58.2\textsuperscript{*}   \\
		BERT\textsubscript{large}\cite{devlin-etal-2019-bert}  & 66.0\textsuperscript{*} & 66.8\textsuperscript{*}    \\
		XLNet\cite{DBLP:conf/nips/YangDYCSL19}   & \textsuperscript{*} & 72.0\textsuperscript{*}  \\
		RoBERTa\textsubscript{large}\cite{DBLP:journals/corr/abs-1907-11692}   & 85.4\textsuperscript{*} & 85.0\textsuperscript{*}  \\
		RoBERTa\textsubscript{large}+MMM\cite{jin2019mmm}   & 88.0\textsuperscript{*} & 88.9\textsuperscript{*}  \\
		\hline
		Albert\textsubscript{xxlarge}\cite{Lan2020ALBERT:}   & 89.2\textsuperscript{$\dagger$} & 88.5\textsuperscript{$\dagger$}  \\
		
		Albert\textsubscript{xxlarge}+DUMA\cite{zhu2020dual}   & 89.3\textsuperscript{$\dagger$} & 90.4\textsuperscript{$\dagger$}  \\
		\hline
		Albert\textsubscript{xxlarge}+DUMA(our implementation)   & 90.7 & 88.6  \\
		Our model (above model with multi-task learning)   & 91.9 & 91.8  \\
		
		\hline
	\end{tabular}
	\caption[caption1]{Results on DREAM dataset, the numbers marked with $\star$ are from ~\cite{jin2019mmm}, the numbers marked with  $\dagger$ are from \cite{zhu2020dual} }
	\label{table_all_result}
\end{table}

\section{Acknowledgement}
The author would like to thank Luis Lastras, Sachindra Joshi, and Chulaka Gunasekara for helpful discussions.

\bibliographystyle{named}
\bibliography{Feb2020_submission}

\end{document}